\pdfoutput=1

\documentclass[11pt]{article}
\usepackage{titling}

\usepackage{acl}

\usepackage{times}
\usepackage{latexsym}
\usepackage{graphicx}
\usepackage{dialogue}

\usepackage[T1]{fontenc}

\usepackage[utf8]{inputenc}

\usepackage{microtype}

\renewcommand\arraystretch{1.5}
\usepackage{array}
\usepackage{multirow}

\title{Unraveling ChatGPT: A Critical Analysis of AI-Generated Goal-Oriented Dialogues and Annotations}

\author{
  Tiziano Labruna\textsuperscript{1}\textsuperscript{2},
  Sofia Brenna\textsuperscript{1}\textsuperscript{2},
  Andrea Zaninello\textsuperscript{1}\textsuperscript{2},
  Bernardo Magnini\textsuperscript{1}\\
  \textsuperscript{1}Fondazione Bruno Kessler - Trento, Italy \\
  \textsuperscript{2}Free University of Bozen-Bolzano, Italy
}

\begin{document}
\maketitle
\begin{abstract}
Large pre-trained language models have exhibited unprecedented capabilities in producing high-quality text via prompting techniques. This fact introduces new possibilities for data collection and annotation, particularly in situations where such data is scarce, complex to gather, expensive, or even sensitive. In this paper, we explore the potential of these models to generate and annotate goal-oriented dialogues, and conduct an in-depth analysis to evaluate their quality. Our experiments employ ChatGPT, and encompass three categories of goal-oriented dialogues (task-oriented, collaborative, and explanatory), two generation modes (interactive and one-shot), and two languages (English and Italian). Based on extensive human-based evaluations, we demonstrate that the quality of generated dialogues and annotations is on par with those generated by humans.
\end{abstract}

\section{Introduction}

Since its initial release in November 2022, ChatGPT has been tested on various tasks, including traditional NLP tasks, domain-specific knowledge such as medicine, fixing computer programs, and even solving neuro-psychological tests used for humans. Despite its high potential, little attention has been paid to ChatGPT's ability to produce annotated data for training purposes \cite{gpt-as-annotator-2022, Huang2023IsCB-ChatGPT-Annotator}. In this paper, we specifically aim to evaluate ChatGPT's ability to generate novel, human-like dialogues and annotate them according to a predetermined taxonomy, which is a very challenging task in NLP.\\
Although we anticipate further improvements in the language model's conversational abilities, generating high-quality training data remains essential for developing adaptable dialogue systems for various domains and conversational contexts. However, collecting human-like dialogues often requires complex and expensive settings to simulate ecological data. Current methodologies include the "Wizard of Oz" method \cite{wizardofoz-reference}, in which a human (i.e., the Wizard) simulates the system's output, the Map Task \cite{Hcrc_map1991} method, initially proposed for an instruction-giving task involving two participants who must collaborate to reproduce the itinerary on a plain map, and transcribing audio and video recordings \cite{wachsmuth-alshomary-2022-mama}.
\\
Dialogue annotation is crucial for training and evaluating dialogue models \cite{budzianowski2018multiwoz}. However, this is a complex and costly process, as annotation schemas are often not standardized and human annotation is time-consuming and error-prone. As a result, only a few annotated dialogue datasets are available, which cover limited dialogue types and domains, with insufficient data and ongoing debates regarding their quality (as exemplified by the various versions of the MultiWOZ dataset). Therefore, simplifying the dialogue collection and annotation process would greatly benefit research in this area.
\\
Meanwhile, very large pre-trained language models (LPLMs), such as BERT \cite{devlin2018bert}, T5 \cite{raffel-t5}, GPT-3 \cite{GPT-3-reference}, LaMDA \cite{lamda-paper}, PaLM \cite{chowdhery2022palm}, and IstructGPT \cite{instruct-gpt}, have demonstrated unparalleled ability to generate high-quality text via prompting strategies \cite{prompting-survey}. While recent studies have examined the potential of these language models as annotators \cite{gpt-as-annotator-2022}, this area remains largely unexplored, especially regarding dialogue generation and annotation.
\\
This paper investigates the ability of Language-based Pre-trained Language Models (LPLMs) to generate and annotate different types of dialogues. We conduct a thorough qualitative analysis of both the generated dialogues and their semantic annotations to evaluate their usefulness in training dialogue systems. Specifically, we consider three dialogue scenarios of varying complexity: (i) task-oriented dialogues \cite{McTear2020} that present a user with a specific need (e.g., booking a restaurant) and an agent that helps the user achieve their goal through dialogue; (ii) cooperative dialogues where two participants collaborate to achieve a shared goal (e.g., matching job requests and offers); and (iii) explanatory dialogues aimed at providing clarifications and explanations on a specific topic through dialogue (e.g., a medical doctor explaining a diagnosis to a patient or a teacher-student scenario).
\\
To evaluate the quality of the generated dialogues, we conducted a crowd-sourced evaluation using a questionnaire based on a 6-point Likert scale, comparing both the dialogues generated by ChatGPT and corresponding human-generated dialogues. Additionally, we used standard measures for dialogue state tracking \cite{DSTCReview}, i.e., Slot Accuracy and Joint Goal Accuracy, to evaluate the quality of the annotations performed by ChatGPT.

The contributions of the paper are as follows:
(ii) We report on the first experiment, to the best of our knowledge,  aiming at generating complex, human-like dialogues through controlled prompting of a large language model (ChatGPT). (ii) We show that the quality of automatically generated dialogues is comparable to that of reference human-generated dialogues. We also show that the quality of the annotations generated by ChatGPT is comparable to that of human annotations. (iii) We report a number of critical limitations, including the generation of hallucinations, which need to be considered in future research.

\section{Background and Related Work}
\label{sec:Background}
In our experiments, we address three types of dialogues: (i) task-oriented dialogues, (ii) collaborative dialogues, and (iii) explanatory dialogues. As for the language model, for all our experiments we use ChatGPT, a recent conversational chatbot built upon a state-of-the-art GPT model.

\subsection{Task-oriented dialogues.}
The objective of task-oriented dialogues is to obtain specific pieces of information that meet the user's requirements, such as booking a restaurant, learning how to open a bank account, or checking tomorrow's weather.
A typical task-oriented dialogue presumes that the user has a definite objective in mind, which is then identified by an operator during the conversation. The operator may pose questions to the user to narrow down the search space and pinpoint the objects that match the user's objectives.
In some situations, the obtained information may be in the form of text, as is the case with Frequently Asked Questions (FAQ) retrieval \cite{DBLP:conf/clic-it/FonsecaMFQM16}. An example of a typical FAQ scenario is a call center, where a customer inquires about a specific procedure, and the operator has to choose a response from a repository of FAQs.
In other scenarios, the information obtained may consist of objects, such as movies or restaurants, that have been extracted from a knowledge base. This is seen in interactive question answering \cite{FERRANDEZ-qallme} or conversational search \cite{DBLP:conf/chiir/RadlinskiC17}. 
\paragraph{MultiWOZ 2.4.} MultiWOZ \cite{budzianowski2018multiwoz} has become one of the most widely used task-oriented conversational datasets. It consists of over 10,000 dialogues spanning seven different domains, including restaurant reservations and train ticket purchases. The dataset was collected using the Wizard of Oz technique, in which one person acted as the user, receiving instructions on the task to be completed, while the other acted as the system, with access to the domain ontology to provide accurate information. Subsequent versions of the dataset brought incremental improvements, culminating in version 2.4 \cite{ye2021multiwoz}, which we use in our experiments.
\\
Semantic annotations in MultiWOZ follow a commonly used schema for training dialogue understanding \cite{louvan2020recent} and dialogue state tracking \cite{balaraman2021recent} components. Annotations are based on triplets consisting of the \textit{domain} (e.g., \textsc{Restaurant}), \textit{slot} (e.g., \textsc{Food}), and \textit{slot-value} (e.g., \textsc{Italian}), which are consistent with the ontology of the conversational domain. These annotations are incremental, meaning that new slot-values mentioned in the dialogue are added to previous ones, enabling to maintain a full belief state of the user's requirements at each step of the conversation. For an example of a MultiWOZ dialogue and its annotations, refer to Appendix \ref{sec:appendix-multiwoz-example}.

\subsection{Collaborative Dialogues}
Collaborative dialogues require the cooperation of the interlocutors to follow the execution of a shared plan. For instance, when a user is planning a vacation with a travel operator, multiple steps must be taken, such as determining the vacation's time and location, reviewing various travel agent proposals, making decisions, and reserving transportation and lodging. Numerous command-based dialogues, in which one participant requests the other to complete a task (e.g., in a car, at home), with varying degrees of complexity, belong to the category of collaborative dialogues and are now prevalent in many personal assistants. 

\paragraph{JILDA.} 
The JILDA Corpus \cite{SUCAMELI-JILDA-2021} is a collection of 525 Italian dialogues between two humans in the job search and offer domain. The dialogues were gathered using the role-taking method in a two-party online chat, where an Applicant seeks assistance from a Navigator, who is a job consultant. The experiment was conducted using an adaptation of the Map Task methodology \cite{Hcrc_map1991}.

Compared to dialogues collected using a crowd-sourcing template-based approach, the JILDA dataset has been shown to exhibit greater lexical variety, syntactic complexity, conversational naturalness, and overall length. In JILDA, similarly to the MultiWOZ dataset, semantic annotations are based on dialogue acts and slot-value pairs associated with each dialogue utterance. These annotations are specific to the conversational domain. An example of a JILDA dialogue can be found in Appendix \ref{sec:appendix-jilda-example}.

\subsection{Explanatory dialogues} 
There is nowadays widespread agreement that explaining a concept is primarily a social and collaborative practice, which is often likened to a dialogue where both the explainer, typically an expert in the field, and the explainee work together to construct an understanding of a particular topic \cite{CAWSEY198969,moore-1994,rohlfing2021explanation}. An example of such an explanatory dialogue can be observed between a patient and a medical doctor. Here, the doctor must explain the reasons for a particular diagnosis, while the patient can seek clarifications and additional information.
However, despite the growing interest in such data, explanatory dialogue datasets are still rare.

\paragraph{WIRED.}
One of the few resources that cater to explanatory dialogues, which we used in our experiments, is the \textit{WIRED 5 Levels} corpus by \citet{wachsmuth-alshomary-2022-mama}. This corpus comprises transcriptions of dialogues from the WIRED video series 5 Levels, all in English, where a field expert (a university teacher) explains over 13 topics (from music harmony to machine learning) to five explainees with varying levels of proficiency (from children to colleagues). The corpus contains 65 dialogues, with a total of 1550 dialogue turns. Furthermore, each turn was manually labeled for the topic, dialogue act, and the type of explanation given. In our experiments, we used the 5-level dialogues (child, teenager, undergrad, postgrad, colleague) that deal with the topic of "machine learning." Refer to Appendix \ref{sec:appendix-WIRED-example} for an example of a WIRED dialogue.

\subsection{Pre-trained LMs and Prompt-Learning}

In recent years, pre-trained language models have been the focus of extensive research, due to their ability to learn from large amounts of data in a self-supervised fashion and achieve impressive results in various tasks \cite{howard2018universal, radford2019language}. One of the latest trends in utilizing these language models is the development of prompt-based techniques, where a textual prompt is given to the model as input to generate the desired output. Such techniques have shown to be highly effective, especially for tasks that require specific outputs and have the advantage of (i) not requiring any parameter updates in the model; (ii) be human readable, and (iii) not requiring in-domain data, unlike fine-tuning techniques. An example of such a model is GPT-3 \cite{GPT-3-reference}, a pre-trained language model that uses the Transformer architecture and an attention mechanism to generate natural language text. For an extensive survey on prompt-based techniques, refer to \cite{Liu2021PretrainPA}.

\paragraph{ChatGPT.}
Given our focus on dialogue, for all our experiments we leverage the interactive interface of ChatGPT \cite{openai2022chatgpt}. ChatGPT is part of the InstructGPT family \cite{ouyang2022training}, which is based on the GPT-3 language model \cite{GPT-3-reference}. Unlike standard GPT-3.5 models, InstructGPT models are optimized for interactive use and can "learn" from their mistakes, making them more aligned with user requests. This is accomplished by creating a reward function using human feedback and using Reinforcement Learning from Human Feedback (RLHF) \cite{christiano2017deep} to optimize it. To train the model, human trainers engaged in conversations as both a user and an AI assistant. The model was fine-tuned through supervised training, and the AI-generated suggestions were given to human trainers for producing their responses. The resulting dataset was combined with the InstructGPT dataset and transformed into a dialogue format.
To collect this data, conversations were conducted between the AI trainers and the model. The trainers ranked the automatically generated messages sampled from several alternative completions, and these reward models were used to fine-tune the model using Proximal Policy Optimization in several iterations.


\paragraph{LMs as Annotators.} The study by \cite{gpt-as-annotator-2022} explored the potential of using GPT-3 \cite{GPT-3-reference} as a data annotator for NLP tasks. Specifically, they conducted experiments on two tasks, namely text classification (sentiment analysis) and named entity recognition using the CrossNER dataset. Their findings demonstrated that employing GPT-3 annotations can achieve comparable results to those obtained from human-labeled datasets, at a 5\% cost.
While this aligns with our research objectives, our intuition is that ChatGPT's prompt-learning performance has significantly improved. As a result, we anticipate that more challenging tasks like dialogue generation and annotation can be addressed.

\section{Methodology}

Our objective is to investigate various aspects related to the generation and annotation of dialogue datasets. Firstly, we aim to evaluate the language model's capability to \textbf{generate} high-quality dialogues that resemble those collected through natural means. Secondly, we intend to assess the model's proficiency in \textbf{annotating} a dialogue based on a predefined annotation scheme. Lastly, we are keen on evaluating the model's capacity to generate and annotate dialogues in languages other than English, particularly in Italian. \\
To achieve this, we propose a three-step methodology:

\begin{itemize}
\item We begin by selecting a dialogue from an existing repository, such as MultiWOZ, as the \emph{reference dialogue}. We then prompt the language model to produce a \emph{generated dialogue} that closely resembles the reference dialogue.
\item Next, we prompt the language model to annotate the generated dialogue using the same annotation schema provided in the existing repository. This results in an \emph{annotated dialogue}.
\item Finally, we evaluate \textit{both} the generated and the reference dialogue using an external evaluation source. This enables us to assess both the quality of the dialogue and its annotation.
\end{itemize}

\subsection{Dialogue Generation}
\label{sec:dialogue_generation}
Here the goal is to create appropriate prompts that can generate a goal-oriented dialogue which is similar in content to a reference dialogue. There are two ways to prompt a language model for a dialogue: provide all the instructions in a single prompt and allow the LPLM to generate the entire dialogue at once (\textit{one-shot} approach), or, instruct the model to simulate the behavior of one participant only (e.g., the system in a task-oriented dialogue) and obtain the dialogue through interaction with a human who plays the role of the other participant (\textit{interactive} approach). For our purposes, which involve collecting dialogues at a low cost, we are more interested in the \textit{one-shot} approach, although the \textit{interactive} approach can provide important feedback on the ability of an LPLM to engage in a conversation with a real user.



\paragraph{Dialogue generation prompting scheme.}
For both \textit{one-shot} and \textit{interactive} approaches, we follow a structured prompt format with the following content areas (more detailed examples are available in the Appendix):

\begin{enumerate}
    \item High level instruction about the task to be executed. E.g., "Create a dialogue between a user and a system." (or in the case of the interactive setting: "Simulate to be a system and respond to a user").

    \item Description of the dialogue context. E.g. "The user asks for information on restaurants in order to make a reservation and the system, called “Cambridge InfoTown” asks the user all the needed information, tells the user whether there is availability for the people, time and date requested and finally makes the reservation, giving a random reference number. Every user message should be followed by a system message. Be polite and don't forget to say goodbye. 
Everything said by "Cambridge InfoTown" needs to be strictly coherent to the Domain Knowledge that I provide you."
    \item Domain knowledge to be used in the dialogue. E.g., "name: pizza hut city centre, address: Regent Street City Centre, area: centre, food: italian, phone: 01223323737, postcode: cb21ab, pricerange: cheap". 
    \item Dialog-specific instructions to be followed by the user (only for the \textit{one-shot} setting). E.g., "For the dialog that you have to generate, the instructions for the user are the following: You are looking for an Indian restaurant in the north [...]".
    \item Instructions on how to  simulate the system (only for the \textit{interactive} setting. E.g., "I will start acting like a user needing of information about restaurants in Cambridge starting from my next message".
\end{enumerate}
Although certain aspects of each dialogue may require customized prompts, we aim to maintain this structure consistent across all our experiments.


\subsection{Dialogue Annotation}
\label{sec:dialogue_annotation}
This section outlines the methodology for annotating a dialogue through prompting the LPLM. We assume that annotations are added to a previously generated dialogue (either by humans or by a LPLM), although there might be the option  where annotations are generated simultaneously with the dialogue generation. Additionally, for the three dialogue types we are considering, we will use the annotation schema as provided in our reference datasets (details in Section \ref{sec:Background}).



\paragraph{Dialogue annotation prompting scheme.}
 The prompt for dialogue annotation follows the following format:

\begin{enumerate}
    \item High level instruction about the task to be executed. E.g., "Write annotations about a dialogue that I will send you".
    \item Instructions on how to produce the annotations. E.g., "Add annotations of the intents and slots for each one of the dialog turns. The annotations have to follow this format:  [...]
The possible intents are: [...]
The possible slots are: [...]
    \item Dialogue to annotate. E.g., "The dialogue that you have to annotate is the following: 1. User: [...] 2. System: [...]".
    
\end{enumerate}

\subsection{Evaluation targets}
\label{se:method-evaluation}
According to the methodology described in section \ref{sec:dialogue_generation}, we have three versions of each dialogue we consider: a \emph{reference dialogue}, a \emph{one-shot generated dialogue} and a \emph{interactive generated dialogue}. Here the goal is to assess the quality of the generated dialogues, in order to establish whether LPLMs are capable of creating dialogues that are comparable to the ones collected by humans.

\begin{table*}[!ht]
\caption{Qualitative assessment for dialogue generation via AMT. Scores assigned through a 6-question questionnaire in a [-3 +3] Likert scale. Each question with 10 responses. $^\dagger$ indicates quality scores that are not statistically different with respect to the correspondent reference quality score.  *Jilda dialogues are in Italian.}
\label{tab:MTurkResults}
\begin{center}
\def\arraystretch{2}
\begin{tabular}{|c|c|c|c||c|}

      \hline
      \textbf{Dataset}& \textbf{Reference (from dataset)} & \textbf{One-shot generation} & \textbf{Interactive generation} & \textbf{Avg.}\\
      \hline\hline
      \textsc{MultiWOZ} & 1.81 & 1.60$^\dagger$ & 1.83 & \textbf{1.75}\\
      \hline
      \textsc{JILDA*} & 1.01 & 0.81$^\dagger$ & 1.33$^\dagger$ & \textbf{1.05}\\
      \hline
      \textsc{WIRED} & 1.61 & 1.78$^\dagger$ & 1.38$^\dagger$ & \textbf{1.59}\\
      \hline\hline
      \textbf{Avg.} & \textbf{1.48} & \textbf{1.40} & \textbf{1.52} & \textbf{1.46}\\
      \hline
\end{tabular}
 \end{center}
\end{table*}

\paragraph{Dialogue quality.}
\label{sec:diag_quality}
Although automatic metrics like BERT-score \cite{zhang2019bertscore} exist for scoring the similarity of a text against a reference,  such metrics are insufficient, as they do not capture the peculiarities of dialogue. To address the issue, we designed a questionnaire to collect human evaluations through crowdsourcing \cite{garland1991mid, lietz2010research, chyung2017evidence}. The questionnaire, reported below, assesses dialogue quality based on six criteria: consistency and quality of content (criteria 1 and 2), formality (3), politeness (4), naturalness (5), and successfulness (6). \\

\begin{center}
\begin{tabular}{ | p{0.2cm} p{6.5cm} | }
\hline
 1. & Requests and responses are consistent across the dialogue. \\ 
 \hline
 2. & The information that is exchanged by participants in the dialogue is realistic. \\ 
 \hline
 3. & The level of formality shown by the two participants is consistent throughout the dialogue. \\
 \hline
 4. & Participants are respectful towards each other (no offensive/discriminatory language is used). \\
 \hline
 5. & The participants' sentences sound spontaneous and natural in the dialogue. \\
 \hline
 6. & The dialogue comes to a conclusion. \\
 \hline
\end{tabular}
\end{center}
\hfill \break \\
Labelers are presented with each criterion as an affirmative declarative clause and asked to rate their overall agreement on a 6-item Likert verbal scale, ranging from "strongly disagree" to "strongly agree". Our Likert scale replicates the commonly used 7-point bipolar disagreement/agreement scale, with increasing agreement value from left to right. We chose not to provide a neutral middle option, which would allow respondents to avoid committing to a direction in their opinion. However, we provided a text box for optional comments and feedback.

\paragraph{Detecting hallucinations.}
\label{sec:hallucinations}
To evaluate whether a dialogue generated by a LPLM includes hallucinations (i.e., generated text that is "nonsensical or unfaithful to the provided source input" \cite{ji2022survey}), we analyze all system turns that reference the domain knowledge base. We request a domain expert to assess whether the information conveyed in these utterances is accurate or not, assigning a score equal to 0  if any of the information provided is inconsistent with the knowledge base, and 1 if all information is correct. This accuracy metric is then averaged across the entire dialogue (or set of dialogues).

\paragraph{Adherence to instructions.}
\label{sec:instructions}
Here we consider the instructions (i.e., prompts) related to the user goals in the dialogue (e.g. "user is looking for an Italian restaurant") and evaluate whether the generated dialogues comply with the given instructions. We only take into account utterances that are relevant to the instructions provided (e.g., utterances such as "Thank you for your help" are disregarded). For each utterance, a domain expert assigns a score of 0 if any of the information presented in the generated dialogue conflicts with the instructions, and 1 if there is no conflict. The scores for each utterance are then averaged to obtain the overall score for the dialogue (or for the set of dialogues).

\paragraph{Annotation quality.} 
\label{sec:annotations}
In order to assess the quality of generated annotations, we rely on a domain expert that determines whether the annotations for the dialogues are correct. We employ Slot Accuracy ($SA$) and Joint Goal Accuracy ($JGA$) as the quality metrics, which are commonly used in literature for dialogue annotation evaluation \cite{dey2022towards}. To compute these metrics, we ask the experts to define the sets of false negatives ($FN$) and false positives ($FP$) triplets (\textit{domain}, \textit{slot}, \textit{slot-value}), such as "\textsc{Restaurant-Food-Italian}," for each conversational utterance and its corresponding annotation. Then, we compute $SA$ using the following formula:
\[SA = \frac{|S| - |FN| - |FP| + |P \cap Q|}{|S|}\]
where $S$ is the set of unique (\textit{domain}, \textit{slot}) pairs in the dataset, $P$ is the set of unique (\textit{domain}, \textit{slot}) pairs from $FN$, and $Q$ is the set of unique (\textit{domain}, \textit{slot}) pairs from $FP$. Therefore, $P \cap Q$ represents the set of \textit{domain-slot} pairs that are present in both $FN$ and $FP$.
$JGA$ is instead a more strict measure and is computed by assigning a score of 0 to any annotation with at least one false negative or false positive, and 1 to annotations that are 100\% correct. Finally, we take the average of the $SA$ and $JGA$ values for each conversational turn to obtain a score for the entire dialogue (or set of dialogues).

\section{Experiments and Results}
The experimental setup for our qualitative analysis comprises three different types of dialogues, with one dataset per type, a model (ChatGPT) that we utilized for dialogue generation and annotation, a quality assessment procedure for generated dialogues, based on a questionnaire created through Amazon Mechanical Turk (AMT), two expert evaluations to gauge ChatGPT's ability to avoid hallucinations and follow instructions, and a quality assessment method for dialogue annotations.

\begin{table*}
\caption{Qualitative assessment of dialogue annotations: slot accuracy, joint goal accuracy, adherence to domain knowledge and to instructions.  *Jilda dialogues are in Italian.}
\label{tab:InternalResults}
\begin{center}
\begin{tabular}{*{5}{|>{ }c<{}}|}\hline
\multirow{2}{*}{\textbf{Dataset}} &
\multicolumn{2}{|c|}{\textbf{Annotation Quality}} & \multirow{2}{*}{\parbox{2cm}{\centering\textbf{Domain}\\ \textbf{Adherence}}} & \multirow{2}{*}{\parbox{2cm}{\centering\textbf{Instructions}\\ \textbf{Adherence}}} \\\cline{2-3}
& Slot Accuracy & Joint Goal Accuracy &  & \\
\hline\hline
      \textsc{MultiWOZ\_one-shot} & 0.88 & 0.50 & 0.89 & 0.83 \\
      \hline
      \textsc{MultiWOZ\_Interactive} & 0.93 & 0.52 & 0.89 & - \\
      \hline
      \textsc{JILDA\_one-shot*} & 0.71 & 0.03 & 1 & 0.92 \\
      \hline
      \textsc{JILDA\_Interactive*} & 0.73 & 0.05 & 0.82 & - \\
      \hline

\end{tabular}
\end{center}
\end{table*}

\subsection{Quality of Generated Dialogues} 
We selected five reference dialogues at random from each of the considered datasets: MultiWOZ, JILDA, and WIRED. Using the \textit{one-shot} and \textit{interactive} dialogue generation methods described in section \ref{sec:dialogue_generation}, we prompted ChatGPT to produce two additional versions of each selected dialogue, resulting in 10 generated dialogues per dataset and a total of 45 dialogues including the reference dialogues. We asked 10 AMT workers to evaluate the overall quality of each dialogue using the questionnaire described in section \ref{se:method-evaluation}. For rating the JILDA dialogues, we created an Italian version of the questionnaire.
\\
We provide additional details in Appendix \ref{sec:appendix-questionnaire}, which includes an example of how evaluators were presented with the task. To facilitate results analysis, we converted the verbal scale to numerical values ranging from -3 (corresponding to "strongly disagree") to +3 (corresponding to "strongly agree").
\\
Table \ref{tab:MTurkResults} presents the average quality scores for the three datasets, including the reference dialogue and the two versions generated by ChatGPT (one-shot generation and interactive generation). Notably, the quality of both human dialogues (score=1.48) and dialogues fully generated by ChatGPT (score=1.40) was perceived by our annotators as substantially equivalent. This result was consistent across the six dimensions considered in the questionnaire. Additionally, the quality of dialogues generated through interaction between ChatGPT and a human (score=1.52) was substantially equivalent to that of human-generated dialogues.
\\
We conducted a test to determine whether the difference in quality between the one-shot and interactive generations was statistically significant with respect to the corresponding reference dialogues. Results, reported in Table \ref{tab:MTurkResults}, showed that there was no significant difference in quality score for almost all of generated dialogues, confirming ChatGPT's capacity to generate high-quality dialogues for the three datasets.
\\
As expected, task-oriented dialogues in MultiWOZ achieved higher quality than those in WIRED, likely due to the simpler structure of the dialogues (e.g., fewer turns). We also observed that the quality of JILDA dialogues consistently scored lower than that of MultiWOZ and WIRED. Upon inspection of the data collected through AMT, we found that this was likely due to labelers' lesser proficiency with Italian, which affected their judgments. 

\subsection{Assessing Hallucinations}
Following the methodology described in section \ref{sec:hallucinations}, we assessed the extent to which ChatGPT generates dialogues with hallucinating characteristics. Since the presence of hallucinations is judged by comparison against domain knowledge, we did not conduct the evaluation on the WIRED dataset, as it does not come with a knowledge base. For MultiWOZ and JILDA we considered both the one-shot and interactive versions (see section \ref{sec:dialogue_generation}). Results, presented in Table \ref{tab:InternalResults} under the column "Domain Adherence", show a relatively high adherence to domain information, with only a few instances of hallucinations.   In the MultiWOZ one-shot experiment, the user requested a Chinese restaurant, but the system suggested a Modern European one, while in the interactive setting, ChatGPT mentioned two restaurants that were not included in the domain knowledge. Overall, ChatGPT remains adherent to the provided knowledge base in most of the cases, although in a few cases (approximately three out of every 20 utterances), it returns responses that are significantly incorrect with respect to the domain knowledge.

\subsection{Evaluating Instruction Adherence}
We followed the methodology described in section \ref{sec:instructions} to measure ChatGPT's ability to adhere to prompt instructions related to the user goals during the dialogue. The WIRED dataset could not be used for this evaluation, as it does not provide specific instructions. Similarly, we could not consider dialogues generated through the interactive method, as the user's turns are not generated by ChatGPT. The results for the one-shot dialogues for both MultiWOZ and JILDA are shown in Table \ref{tab:InternalResults} under the column "Instruction Adherence".
The evaluation demonstrated satisfactory results in ChatGPT's ability to adhere to instructions related to the user goals. However, there were cases where it made errors. For example, in JILDA, one of the generated dialogues had the user claiming to possess certain skills that were not mentioned in the instructions. In another case, in MultiWOZ, the user made a reservation for a date and time that did not correspond to what was specified in the instructions. 
Similar to adherence to domain knowledge, ChatGPT generally correctly follows given instructions, but, in the rare cases where this does not happen, generated the errors are critical.

\subsection{Quality of Generated Annotations}
 To assess ChatGPT's ability to generate proper annotations we used Slot Accuracy and Joint Goal Accuracy as metrics, as described in section \ref{sec:annotations}. The WIRED dataset could not be used as ChatGPT was not able to produce complete annotations, likely due to the length of the dialogues of this dataset.  
 Table \ref{tab:InternalResults}, column "Annotation Quality", shows the results of generated annotations for both MultiWOZ and JILDA, both for the one-shot and interactive versions, which were previously generated by ChatGPT  (section \ref{sec:dialogue_generation}). The results for the MultiWOZ dialogues are significantly high, with performances comparable to those achieved by recent DST systems. On the JILDA datasets, instead, the JGA is considerably lower. This discrepancy may be due to the longer length of the JILDA's dialogues, greater lexical variety and syntactic complexity, and the more intricate annotation schema.

 \begin{figure}
\centering
\includegraphics[width=220pt]{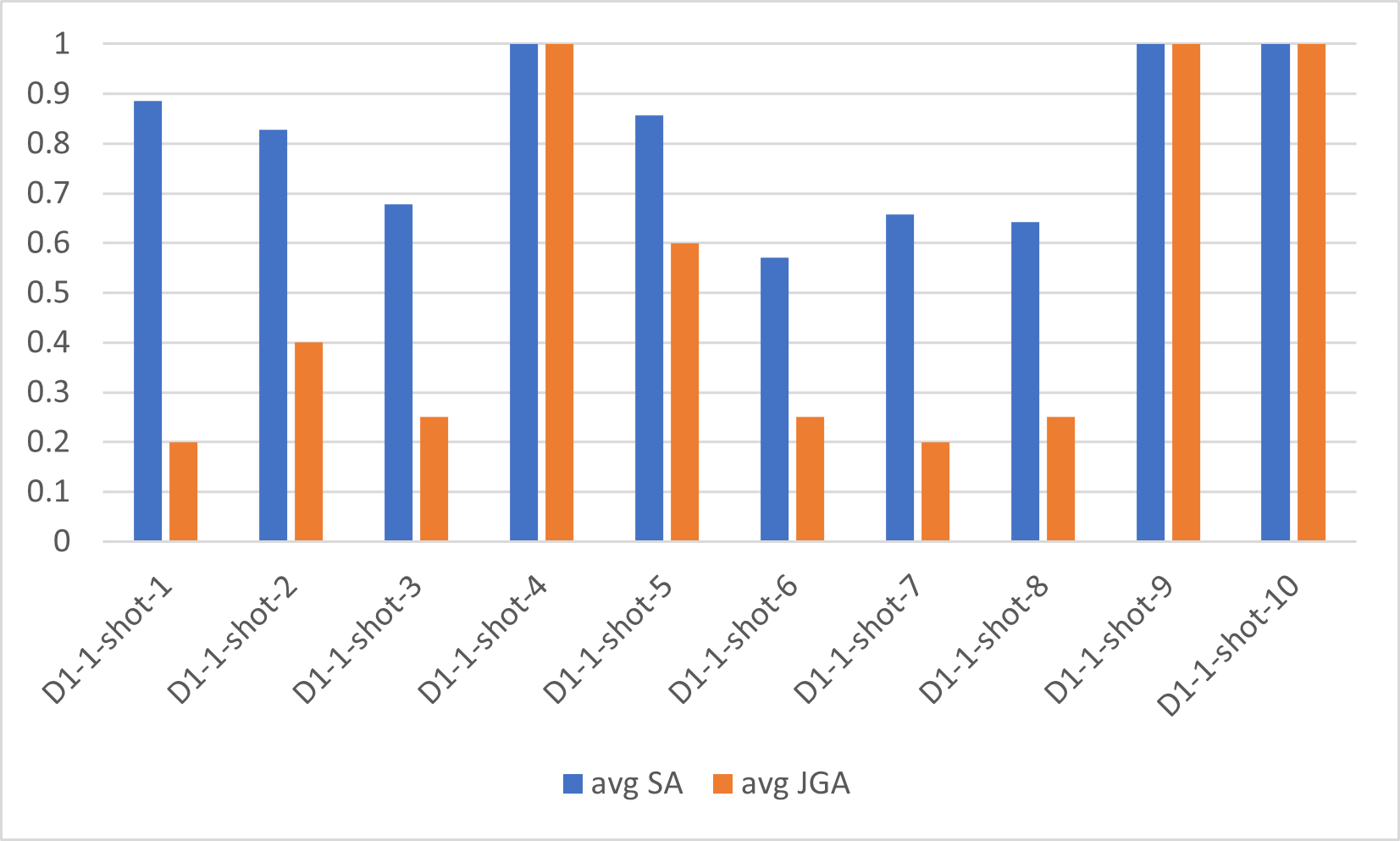}
\caption{ChatGPT's stability. Each pair of columns represents the Slot Accuracy and Joint Goal Accuracy of generated annotations for 10 different runs of the same prompt.}
\label{fig:stochastic}
\end{figure}

\subsection{ChatGPT Stability}
We conducted an assessment of the stability of the ChatGPT output in a series of runs. We used a \textit{one-shot} prompt from the MultiWOZ domain, sent in the exact same way for 10 different times. While all the 10 generated dialogues did not present any hallucination and they were all adherent to the given instructions, we noted several differences in the generated annotations. Figure \ref{fig:stochastic} shows the scores for $SA$ and $JGA$ for each one of the runs. As it can be noted, only 3 out of the 10 dialogues presented perfect annotations. In general, the quality of the annotations was very inconsistent,  demonstrating that ChatGPT, at least the version currently available through interface, is  lacking determinism.

\section{Conclusion}

In this paper, we presented an extensive evaluation conducted by humans, aimed at assessing the quality of dialogues that were generated and then annotated by ChatGPT. This is, to be the best of our knowledge, the first time that a quality analysis has been conducted on fully automatically generated dialogues. The experimental setup included three types of dialogues (task-oriented, collaborative, and explanation-based), two generation methods (interactive and one-shot), and two languages (English and Italian). Our findings demonstrate that the quality of both the generated dialogues and annotations is comparable to the corresponding reference dialogues found in available datasets. While ChatGPT has shown surprising capacity to manage complex prompts, it has also shown a number of limitations. Two of them are particularly critical for practical applications: generated annotations contain a small number of hallucinations, which may disturb the use of such data for training purposes, and, such hallucinations tend to increase as we prompt ChatGPT with the same prompt.





\bibliographystyle{acl_natbib}
\bibliography{acl}

\appendix

\section{Appendix: Dialogue Examples}
\label{sec:appendix-dialogue-examples}

\subsection{MultiWOZ Example} 
\label{sec:appendix-multiwoz-example}
\begin{dialogue}
\speak{User} "I am looking for an expensive Italian restaurant.\end{dialogue}
\textit{metadata}: \{"restaurant: \{"pricerange": "expensive", "food": "Italian"\}\}
\begin{dialogue}
\speak{System} "There is an expensive Italian restaurant named Frankie and Bennys at Cambridge Leisure Park Clifton Way Cherry Hinton. Would you like to go there or choose another?"

\speak{User} "Great yeah that sounds great can you book a table for 5 people at 11:30 on Sunday?"
\end{dialogue}
\textit{metadata}: \{"restaurant: \{"pricerange": "expensive", "food": "Italian", "people": "5", "time": "11:30", "day": "Sunday"\}\}
\begin{dialogue}
    
\speak{System} "Unfortunately there are no tables available, please try another day or time slot."

\speak{User} "How about 10:30 on sunday ? And may I have a reference number?"
\end{dialogue}
\textit{metadata}: \{"restaurant: \{"pricerange": "expensive", "food": "Italian", "people": "5", "time": "10:30", "day": "Sunday"\}\}
\begin{dialogue}
\speak{System} "I was able to book a table for 5 at 10:30 on Sunday. Your reference number is MBC9E6AL. Is there anything else I can help you with today?"

\speak{User} "No, that is all. Thank you so much for your help."
\end{dialogue}
\textit{metadata}: \{"restaurant: \{"pricerange": "expensive", "food": "Italian", "people": "5", "time": "11:30", "day": "Sunday"\}\}
\begin{dialogue}
\speak{System} "You're welcome . Good - bye."
\end{dialogue}

\subsection{JILDA Example}
\label{sec:appendix-jilda-example}
\begin{dialogue}
\speak{System} "Buongiorno"
\speak{User} "Buongiorno!"
\speak{System} "Ciao, sono Giulia! Oggi sono qui per farti da Navigator. Di cosa sei alla ricerca? :)"
\speak{User} "Ciao! Sto cercando lavoro come ricercatore."
\speak{System} "Interessante! Dimmi, in quale settore sei specializzato?"
\speak{User} "Sono specializzata in informatica e robotica, vorrei trovare lavoro in questi settori se possibile."
\speak{System} "Sei alla ricerca di uno stage o preferiresti una posizione lavorativa più stabile?"
\speak{User} "Preferirei una posizione più stabile, a tempo determinato o indeterminato."
\speak{System} "Molto bene. Dimmi, hai già esperienze lavorative nel settore?"
\speak{User} "Purtroppo non ho esperienze in azienda, ma ho un dottorato in Ingegneria Robotica."
\speak{System} "Perfetto! Puoi dirmi quali sono i linguaggi di programmazione e i software con cui hai più confidenza?"
\speak{User} "Certo! So programmare molto bene in C++, java e PhP."
\speak{System} "Ok, hai conoscenze di javascript, MySQL e bash Linux?"
\speak{User} "Li conosco molto superficialmente."
\speak{System} "Ok, dalle informazioni che mi hai dato ho trovato un lavoro che potrebbe interessarti. Un'azienda estera sta cercando una persona che ricopra il ruolo di software engineer. L'annuncio specifica che preferirebbero una persona con esperienze lavorative pregresse in questo settore, saresti comunque interessato a candidarti?"
\speak{User} "Sì, potrebbe interessarmi. Di che cosa si occupa l'azienda?"
\speak{System} "Non viene specificato. Tuttavia il ruolo che dovresti ricoprire consisterebbe nello sviluppare un nuovo progetto in tutte le sue fasi ( Design + Coding + Test + Packaging + Delivery) oltre ad aggiornare i progetti esistenti. Specificano comunque che il lavoro offre la possibilità ai lavoratori di esprimere la propria creatività rispetto a prodotti usati ogni giorno da milioni di persone."
\speak{User} "Interessante. Come si chiama l'azienda?"
\speak{System} "Il nome dell'azienda è Jobtome, una start up formata da circa 50 persone. Vuoi che ti dia il loro contatto?"
\speak{User} "Sì, grazie. Così poi posso contattarli direttamente io."
\speak{System} "info@azienda.com. Hai bisogno di altre informazioni?"
\speak{User} "Al momento no, grazie."
\speak{System} "Spero di esserti stata d'aiuto! Buona giornata :)"
\speak{User} "Sì, sei stata molto gentile! Buona giornata anche a te :)"
\end{dialogue}

\subsection{WIRED Example}
\label{sec:appendix-WIRED-example}
\begin{dialogue}
\speak{Expert} Hi.
\speak{Child} Hi.
\speak{Expert} I'm Hilary, what's your name?
\speak{Child} I'm Brynn.
\speak{Expert} Do you know what machine learning means? Have you heard that before?
\speak{Child} No.
\speak{Expert} So machine learning is a way that we teach computers to learn things about the world by looking at patterns and looking at examples of things. So can I show you an example of how a machine might learn something?
\speak{Child} Sure.
\speak{Expert} [Hilary] So is this a dog or a cat?
\speak{Child} It's a dog.
\speak{Expert} And this one?
\speak{Child} A cat.
\speak{Expert} And what makes a dog, a dog and a cat, a cat?
\speak{Child} Well, dogs are very playful, I think, more than cats. Cats lick themselves more than dogs, I think.
\speak{Expert} That's true. Do you think, if we look at these pictures, do you think maybe we could say, Well, they both have pointy ears, but the dogs have a different kind of body and the cats like to stand up a little different.? Do you think that makes sense?
\speak{Child} Yeah. Yeah.
\speak{Expert} What about this one?
\speak{Child} A dog. A cat. I think, a cat? Because it's more skinny. And also, its legs are like really tall and its ears are a little pointy.
\speak{Expert} This one's a jackal. And it's actually a kind of dog. But you made a good guess. That's what machines do too. They make guesses. Is this a cat or a dog?
\speak{Child} [Brynn] None.
\speak{Expert} [Hilary] None. What is it?
\speak{Child} It's humans.
\speak{Expert} And how did you know that it's not a cat or a dog?
\speak{Child} Because cats and dogs... Because they walk on their paws and their ears are like right here, not right here, and they don't wear watches.
\speak{Expert} And so, you did something pretty amazing there. Because we asked the question, Is it a cat or a dog? And you said, I disagree with your question. It's a human. So machine learning is when we teach machines to make guesses about what things are based on looking at a lot of different examples. And I build products that use machine learning to learn about the world and make guesses about things in the world. When we try to teach machines to recognize things like cats and dogs, it takes a lot of examples. We have to show them tens of thousands or even millions of examples before they can get even close to as good at it as you are. Do you have tests in school?
\speak{Child} Yeah, I have. After every unit, we have a review and then we have a test.
\speak{Expert} Are those like the practice problems you do before the test?
\speak{Child} Well, just like everything that's gonna be on the test is on the review.
\speak{Expert} Which means that in the test, you're not seeing any problems that you don't know how to solve. As long as you did all your practice, right?
\speak{Child} Yeah.
\speak{Expert} So machines work the same way. If you show them a lot of examples and give them practice, they'll learn how to guess. And then when you give them the test, they should be able to do that. So we looked at eight pictures and you were able to answer really quickly. But what would you do if I gave you 10 million examples? Would you be able to do that so quickly?
\speak{Child} No.
\speak{Expert} So one of the differences between people and machines is that people might be a little better at this, but can't look at 10 million different things. So now that we've been talking about machine learning, is this something you want to learn how to do?
\speak{Child} Kind of. Because I kind of want to become a spy. And we used to do coding, so I may be kind of good at it.
\speak{Expert} And machine learning is a great way to use all those math skills, all those coding skills, and would be a super cool tool for a spy.
\end{dialogue}

\section{Appendix: Questionnaire used to assess dialogue quality: statements and verbal tags.}
\label{sec:appendix-questionnaire}

\subsection{English version}
Statements:
\begin{enumerate}
    \item Requests and responses are consistent across the dialogue.
    \item The information that is exchanged by participants in the dialogue is realistic.
    \item The level of formality shown by the two participants is consistent throughout the dialogue.
    \item Participants are respectful towards each other (no offensive/discriminatory language is used).
    \item The participants' sentences sound spontaneous and natural in the dialogue.
    \item The dialogue comes to a conclusion.
\end{enumerate}
Items:

Strongly disagree, Disagree, Somewhat disagree, Somewhat agree, Agree, Strongly Agree.

\subsection{Italian version}
Affermazioni:
\begin{enumerate}
    \item Le richieste e le risposte nel dialogo sono coerenti.
    \item Le informazioni che i partecipanti si scambiano nel dialogo sono realistiche.
    \item Il livello di formalità dei due partecipanti è mantenuto costante per tutto il dialogo.
    \item I partecipanti sono rispettosi l’uno verso l’altro (non usano linguaggio offensivo o discriminatorio).
    \item Le frasi dei partecipanti sono spontanee e naturali nel contesto del dialogo.
    \item Il dialogo arriva ad una conclusione.
\end{enumerate}
Items:

Fortemente in disaccordo, In disaccordo, Leggermente in disaccordo, Leggermente d'accordo, D'accordo, Fortemente d'accordo.

\section{Appendix: Prompts used for dialogue generation}
\label{{sec:appendix-prompt-generation}}
\subsection{One-shot approach}

\paragraph{JILDA}
Create a dialogue between a user and a system in Italian.
The user asks for information on job offers in order to find a suitable one for himself or herself and the system, called Navigator, asks the user all the needed information, tells the user whether there is availability for the skills, requests of the user and finally matches the job offer, giving the company e-mail address as contact. Every user message should be followed by a system message. Be polite and don't forget to say goodbye. 
Everything said by the system needs to be strictly coherent to the knowledge base that I provide you, and everything said by the user needs to be strictly coherent to the user's CV that I provide you.
\\Knowledge base:\\Job offers:
\begin{enumerate}
    \item job\_description1
    \item job\_description2
    \item job\_description3
    \item job\_description4
    \item job\_description5    
\end{enumerate}
User CV:
\textit{\\user\_cv\_D1}

\paragraph{WIRED} Create a dialogue between an Explainer (a University teacher) and an Explainee.
The Explainer and the Explainee hold a conversation on a topic where the Explainer is an expert. The dialogue is a spoken dialogue. The dialogue is explanatory: this means that each turn has a relation to the main topic (e.g. main topic, subtopic, related topic, ...), performs a dialogue act (e.g. check question, agreeing statement, informing statement, ...), and makes an explanation move (e.g. test understanding, request explanation, provide explanation, ...).

The dialogue needs to comply to the knowledge base that I provide to you.

Knowledge base:
Explainer is a machine learning scientist.
Explainee is a computer scientist named Claudia.
Explainee asks about: the Explainer's view on the democratization of machine learning and avaialbility of tools, open questions in the ethics of machine learning, applications of machine learning in fields like agriculture, 
Explainer asks about: things that are holding back from applying machine learning to fields useful to society
Explainer talks about: democratization and accessibility of tools in machine learning, representativeness of data, reproducibility, future of machine learning
Explainee talks about: transparency in machine learning, biases in data, students nowadays

The dialogue should be friendly, informative, and provide details and examples. It should be around 15 turns, and turns can be somehow long. Don't forget to greet at the beginning.

\subsection{Interactive approach}
\paragraph{JILDA}
Simulate to be a system and respond to a user in Italian.
The user asks for information on job offers in order to find a suitable one for himself or herself and the system, called Navigator, asks the user all the needed information, tells the user whether there is availability for the skills, requests of the user and finally matches the job offer, giving the company e-mail address as contact. Every user message should be followed by a system message. Be polite and don't forget to say goodbye. 
Everything said by the system needs to be strictly coherent to the knowledge base that I provide to you.
\\Knowledge base:\\Job offers:
\begin{enumerate}
    \item job\_description1
    \item job\_description2
    \item job\_description3
    \item job\_description4
    \item job\_description5    
\end{enumerate}
User CV:
\textit{\\user\_cv\_D1}
\\I will start acting like a user needing of information about job offers starting from my next message.

\paragraph{WIRED} Let's play a roleplay. We simulate a dialogue between a system, called the Explainer (pretending to be a University teacher) and an Explainee. 
You play the system (the Explainer), and start with the first turn, and I play the Explainee. Please DO NOT write sentences for both turns: you must ONLY play the Explainer, and start the conversation. Then you wait for me to respond. 

We hold a conversation on a topic where the you are an expert. The dialogue is explanatory: this means that each turn has a relation to the main topic (e.g. main topic, subtopic, related topic, ...), performs a dialogue act (e.g. check question, agreeing statement, informing statement, ...), and makes an explanation move (e.g. test understanding, request explanation, provide explanation, ...).
The dialogue needs to comply to the following information:

Explainer (you): "machine learning scientist".
Explainee (me): "computer scientist" named Claudia.
Possible topics: democratization and accessibility of tools in machine learning, representativeness of data, reproducibility, future of machine learning, applications of machine learning to fields useful to society, transparency in machine learning, biases in data, students nowadays, the democratization of machine learning, availability of tools, ethics of machine learning, applications of machine learning in socially relevant fields.

\section{Appendix: Prompts used for dialogue annotation}
\label{{sec:appendix-prompt-annotation}}
\paragraph{JILDA}
Write annotations about a dialogue that I will send you. The annotations have to follow this format: 
metadata: \{slot\_name1: slot\_value1, slot\_name2: slot\_value2, slot\_value3, slot\_name3, ...\}
\\The possible slots are: "job\_description" (description of the job offered), "contract" (type of contract of the job offer), "duties" (the main duties of the job position), "skills" (skills requested for the job), "past\_experience" (past experience required for the job), "degree" (degree or qualification required for the job), "age" (age required for the job), "language" (knowledge of foreign languages), "area" (the field where the company operates), "company\_name" (the name of the company), "company\_size" (the number of workers in the company), "location" (the place where the company is).
\\The annotations are cumulative and need to keep track of all information that has been provided or selected only by the User until that moment of the conversation. 
\\Example of dialog: \textit{\\example\_dialogue}
\\Example of annotations: \textit{\\example\_annotation}
\\The dialogue that you have to annotate is the following: \textit{\\dialogue\_D1}
\\Annotate only the User's turns.

\end{document}